\documentclass[10pt,leqno]{amsart}
\usepackage[numbers]{natbib}
\usepackage{graphicx}
\usepackage{indentfirst,csquotes}

\usepackage{amssymb,amsthm,amsmath}
\usepackage{xcolor,paralist,hyperref,fancyhdr,etoolbox}

\hypersetup{ colorlinks=true, linkcolor=black, filecolor=black, urlcolor=black }

\usepackage{lipsum}

\usepackage{times}
\usepackage{soul}
\usepackage{url}

\usepackage[section]{placeins}
\usepackage{subcaption}
\usepackage{diagbox}
\usepackage{lscape}

\hypersetup{colorlinks=true,
	linkcolor=cyan,
	anchorcolor=cyan,
	citecolor=cyan}

\usepackage{mathtools}
\usepackage{bbding}
\usepackage{pifont}

\usepackage{ragged2e} 
\usepackage{makecell, multirow, tabularx}
\usepackage{color}
\usepackage{float}

\usepackage{wrapfig}
\usepackage{bm}
\begin{document}
	\title{A multi-scale feature enhanced graph neural network for fluid dynamics prediction in complex geometries} 
\author[Li et al.]{Li Xiao, Tianyu Li, Yiye Zou, Mingjie Zhang, and Xiaogangd Deng}
	\date{\today}
	
	\begin{abstract}
		
		Industrial design in fields such as vehicle and aerospace engineering often relies on large-scale numerical simulations to evaluate fluid dynamics performance, which can incur substantial computational costs. Deep neural networks have shown promise in improving simulation efficiency, especially graph neural networks (GNNs), which demonstrate great potential due to their flexibility with unstructured data. However, GNNs face challenges when dealing with tasks involving complex geometries and large-scale meshes. In this paper, we propose the Multi-scale Feature Enhanced Graph Neural Network (ME-GNN) to tackle these challenges. ME-GNN employs a graph neural network with a two-step message-passing mechanism to capture detailed local features effectively. Additionally, it integrates an Attention U-Net with uniform grid discretization, enabling the extraction of both fine and coarse features. The model also utilizes K-hop sampling to construct subgraphs, facilitating efficient training on large datasets while preserving detailed local features. We evaluated ME-GNN on three benchmark datasets and achieved state-of-the-art results: a relative \(L_2\) error of 0.0196 for the velocity field and 0.0556 for the surface pressure on ShapeNet-Car, a normalized mean squared error of 0.0033 for the flow field on AirfRANS, and a relative \(L_2\) error of 0.1416 for the surface pressure on DrivAerNet.
	\end{abstract} 
	
	\maketitle

	\begingroup
	\renewcommand{\thefootnote}{\fnsymbol{footnote}}
	\footnotetext[1]{Li Xiao and Tianyu Li contributed equally to this work.}
	\footnotetext[2]{Corresponding author: Mingjie Zhang.}
	\endgroup
	
	\bigskip

	\section{Introduction}
	\label{sec:introduction}

	Accurate fluid dynamics prediction is crucial in the automotive and aerospace industries, as it significantly influence reliability, fuel efficiency, and safety. Additionally, it enables performance optimization, environmental sustainability, and innovative design. While traditional Computational Fluid Dynamics (CFD) methods deliver high accuracy, they are computationally expensive, time-intensive, and require specialized expertise. These limitations present significant challenges for iterative design processes and rapid prototyping, where swift fluid dynamics prediction are critical.

	Recent advances in deep learning offer promising alternatives that can significantly reduce computation time while maintaining high accuracy~\cite{li2023geometryinformed,anonymous2023geometryguided}. Deep learning models can approximate complex nonlinear relationships and have demonstrated success in various fields, including computer vision and natural language processing. In the context of CFD, data-driven models can learn from simulation data to predict flow fields and aerodnamic forces\cite{elrefaie2024drivaernet,tran2024aerodynamics,umetani2018learning}, offering a potential solution to the computational bottlenecks of traditional methods. Graph-based neural network architectures offer flexibility for unstructured data, as they have the potential to learn from simulation data represented on a grid. However, developing a graph-based model for fluid dynamic prediction presents several challenges:
	\paragraph{\textbf{Complex geometry and flow structures.}} Geometric surfaces exhibit intricate features across various spatial scales, ranging from smooth to sharp, distinct edges. Complex geometry exhibits features at different spatial scales. For example, a car has a large-scale overall shape and small-scale side mirrors. In addition, fluid flow problems also involve multi-scale phenomena, such as wake flows and vortices of various sizes. This requires neural networks to be able to extract features at different scales.
	
	\paragraph{\textbf{Large Scale Irregular Data.}} In CFD, the datasets used to model fluid dynamics performance are often large and irregular, including high-resolution surface mesh with tons of points and intricate geometries~\cite{elrefaie2024drivaernet}. Standalone neural network architectures, such as Multi-Layer Perceptrons (MLPs) and Convolutional Neural Networks (CNNs), are inherently limited in handling large-scale irregular data due to their reliance on regular grid. While Graph Neural Networks (GNNs) offer flexibility for irregular data, they struggle to model long-term relationships effectively. Similarly, Transformers, despite their powerful modeling capabilities, suffer from high computational costs driven by their quadratic complexity, making them less efficient for large-scale applications.

\begin{wrapfigure}{r}{8cm}
		\centering
		\includegraphics[width=0.5\textwidth]{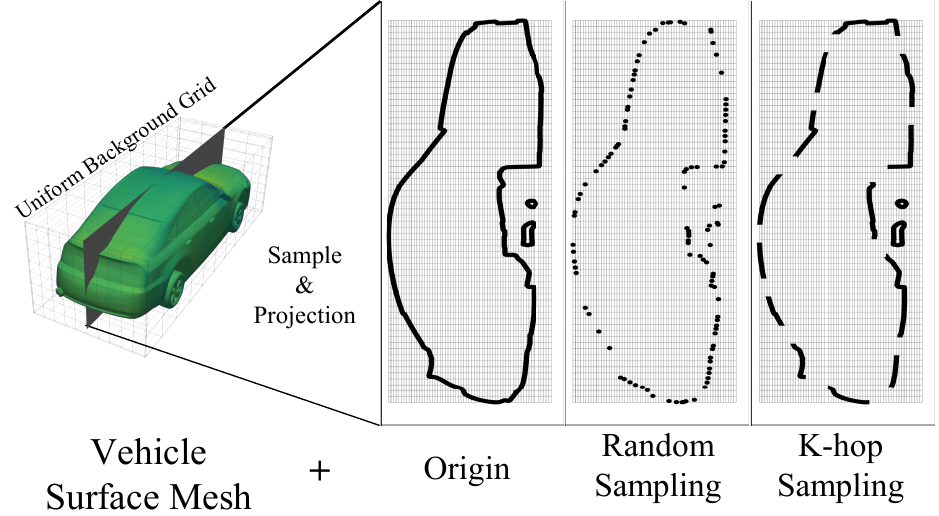}
		\caption{Different sampling for DrivAerNet.}
		\label{fig:bgm_sample}
		\vspace{-0.2cm}
	\end{wrapfigure}

	\paragraph{\textbf{Sampling and Aliasing.}} Due to the large scale of simulation results, sampling is often used during training, which may impact the convergence performance of the neural network. For example, we started by randomly sampling the vehicle and projecting the vertices that are within a certain distance from the central plane, as shown in Fig.\ref{fig:bgm_sample}.  
	
	Compared to the original full mesh, this approach creates incomplete representations of certain parts, especially in the front and underside regions. These gaps may lead to inaccuracies during training, potentially slowing model convergence and reducing overall performance.

\begin{figure}
		\centering
		\includegraphics[width=1.0\textwidth]{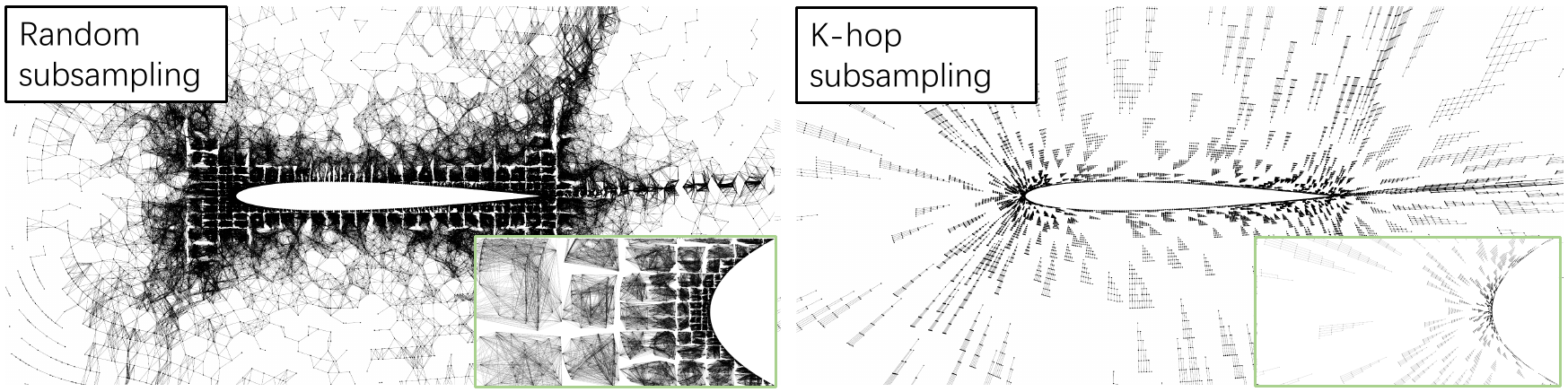}
		\caption{Different sampling for AirfRANS.}
		\label{fig:sub_sampling_2d}
	\end{figure}

	Additionally, as shown in Fig.\ref{fig:sub_sampling_2d}, graph neural networks use random sampling and employ rule-based edge construction methods, such as radius graph or k-nearest neighbors (kNN). Using random sampling methods, there is a good distribution across the global spatial range, but it disrupts the original topological structure in local connections. This randomness during the entire training process can significantly disrupt the performance of the graph neural network. On the other hand, the K-hop~\cite{nikolentzos2020k} sampling method preserves the original local structure, but it leads to sparse sampling in the global space. While K-hop sampling can maintain good local features, its sparse sampling results make it difficult for the graph neural network to establish long-range dependencies.

	To address these challenges, we propose a hybrid deep-learning approach named Multi-scale Feature Enhanced Graph Neural Network (ME-GNN), which utilizes the strengths of CNN-based and graph-based methodologies. Our method combines a Signed Distance Field (SDF) voxel representation or a background grid with an Attention U-Net, along with a K-hop sampled irregular mesh processed through the Finite Volume Graph Network (FVGN)~\cite{li2024predicting, li2024fully}. The K-hop~\cite{nikolentzos2020k} sampling method is particularly effective at preserving detailed local information within subgraphs, providing local details feature compared to random sampling, as illustrated in Fig.\ref{fig:bgm_sample}, such as the frontal region and the underbody area of the vehicle. This approach is analogous to the overset mesh method commonly used in traditional CFD to handle complex geometries. 
	
	The structure of this paper is as follows: Sec.\ref{sec:relwork} reviews related work, establishing the context and motivation for this study. Sec.\ref{sec:Method} describes the proposed method and network architectures. Sec.\ref{sec:exp} presents experimental results and evaluations. Sec.\ref{sec:eccicency_analy} analyzes computational efficiency. Sec.\ref{sec:dis_and_concl} concludes with key contributions and findings.

	\section{Related Work}\label{sec:relwork}
	\subsection{Neural Operator}
	
	Neural Operators are a class of neural networks designed to learn mappings between function spaces. Recent developments include the Deep Operator Network (DeepONet)~\cite{lu2021learning}, which approximates operators using neural networks based on the universal approximation theorem for operators, and the Fourier Neural Operator (FNO)~\cite{li2020fourier}, which leverages the Fourier transform to efficiently parameterize integral operators.
	
	These neural operator methods have demonstrated remarkable performance in approximating solution operators for partial differential equations (PDEs), offering advantages in generalization across different input functions and scalability to high-dimensional problems. Their applications span a wide range of scientific domains, including fluid dynamics~\cite{li2020fourier} and climate modeling~\cite{pathak2022fourcastnet}.
	
	Addressing irregular meshes has recently become a key focus in neural operator models, which have shown notable success. For instance, GNO~\cite[e.g.]{li2020neural} leverages graph neural operators, GINO~\cite{li2023geometryinformed} integrates the strengths of GNO and FNO to extract both frequency-domain and spatial features. Transformer-based approaches have also demonstrated exceptional performance, with GNOT~\cite{hao2023gnot} utilizing transformers and 3D-GeoCA~\cite{anonymous2023geometryguided} incorporating 3D vision pretraining to achieve impressive results. To tackle the computational complexity of transformer self-attention, methods such as Galerkin~\cite{Cao2021ChooseAT} and Transolver~\cite{wu2024transolver} have been proposed, with Transolver achieving remarkable efficiency and performance through learnable slice-based attention.
	
	\subsection{Graph Neural Networks}
	
	GNNs have become a powerful tool for modeling and analyzing data with graph structures~\cite{monti2017geometric,zhou_graph_2020}. In the context of physical simulations and PDEs, GNNs have been employed to model interactions in systems represented by mesh or point clouds~\cite{battaglia2018relational, pfaff2021learning, sanchez2020learning}.
	
	Mesh-based GNNs, such as MGN (MeshGraphNets)~\cite{pfaff2021learning}, use graph representations of physical domains to learn dynamics and have shown promise in fluid simulations. During training, message-passing on a mesh graph can adapt the mesh discretization, which is efficient for learning mesh representation, However, GNNs are often constrained to capturing local features due to computational limitations in message passing over large graphs. This can limit their ability to model long-range dependencies and global features critical in complex physical systems.
	
	\subsection{Geometry-Informed PDEs Learning}
	
	Geometry-Informed PDEs Learning combines geometric processing techniques with deep learning to handle complex shapes and spatial domains. Techniques in this area aim to incorporate geometric priors or representations into neural networks to improve their ability to learn from and generalize across different shapes~\cite{anonymous2023geometryguided,li2023geometryinformed}.
	
	The Signed Distance Function (SDF) has been widely used to implicitly represent geometries. Geometry-Informed Neural Operators (GINO)~\cite{li2023geometryinformed} integrate geometric information directly into network architectures. In the context of PDEs and CFD, Geometry-Adaptive Convolutions~\cite{gao_phygeonet_2021} handle irregular domains by adapting convolutional operations to the underlying geometry.

	\section{Method}\label{sec:Method}

	\subsection{Problem Formulation}\label{ssec:problemform}
	
	We consider a family of problems where the solution is governed by PDEs defined over a spatial domain \( \Omega \subset \mathbb{R}^d \), where \( d \) indicates the spatial dimensions. The domain \( \Omega \) represents the region or surface where boundary conditions are applied.
	
	Our objective is to learn a solution \( \Psi \) that maps from an discretized input function space \( \mathcal{A} \) to the discretized solution space \( \mathcal{H} \). The input \( a \in \mathcal{A} \) encapsulates all necessary information for the PDEs problem, such as boundary conditions, source terms, and any parameters characterizing the system. The solution space \(\mathcal{H}\) represents the potential solutions of the PDEs, which can be defined over the entire domain \( \Omega \): \(\Psi: \mathcal{A} \rightarrow \mathcal{H}\), for any input \( a \in \mathcal{A} \), the solution is given by \( \Psi(a) = u \in \mathcal{H} \).
	
	To approximate the true operator \(\Psi\), we use a parameterized neural network \(\hat{\Psi}_w\) with parameters \( w \in \mathbb{R}^p \). The neural network receives the discretized input \( \mathcal{A}_k \) and produces a prediction: \(\hat{\Psi}_w(\mathcal{A}_k) = \{ \hat{u}_k^i \}_{1 \leqslant i \leqslant N},\) where \(\hat{u}_k^i\) approximates the solution \( u_k \) at discretized points within \(N\) data samples. The learning objective is to minimize the relative \( L_2 \) error between the predicted solution and the ground truth across the training dataset:
	
	\begin{equation}
		\min_{w \in W} \frac{1}{D} \sum_{k=1}^D \frac{\left\| \hat{\Psi}_w (\mathcal{A}_k) - \{ u_k^i \}_{1 \leqslant i \leqslant N} \right\|_2}{\left\| \{ u_k^i \}_{1 \leqslant i \leqslant N} \right\|_2}
	\end{equation}
	
	where:
	\begin{itemize}
		\item \(\hat{\Psi}_w (\mathcal{A}_k)\) is the predicted solution at discretized points.
		\item \(\{ u_k^i \}_{1 \leqslant i \leqslant N}\) is the solution at those points.
		\item \( D \) is the total number of training samples.
		
	\end{itemize}

	\begin{figure}
		\centering
		\includegraphics[width=0.85\textwidth]{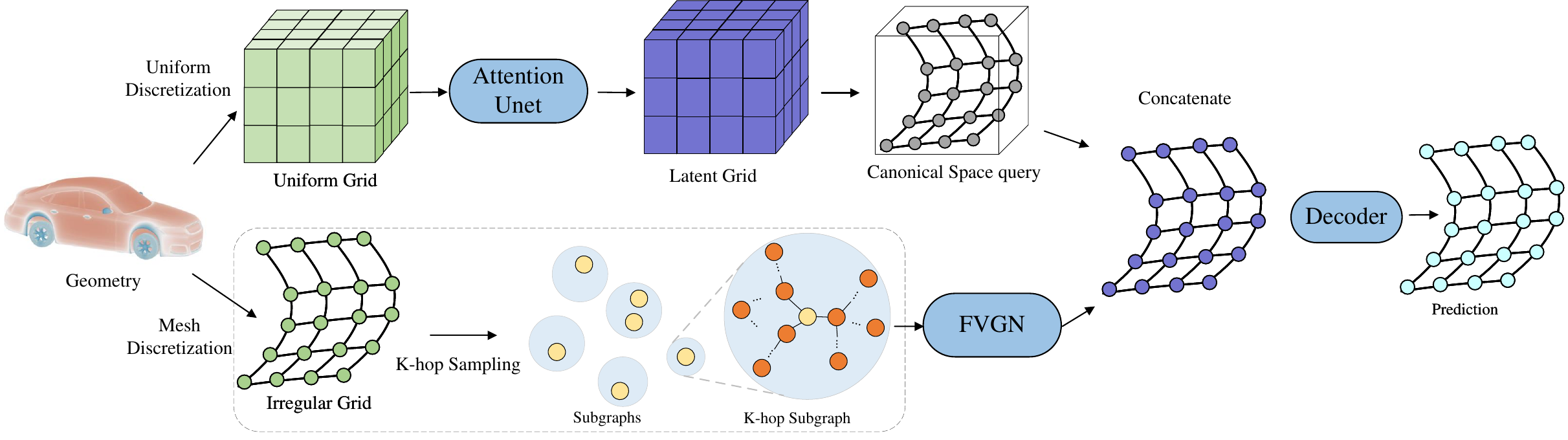}
		\caption{Method overview. The input to ME-GNN comprises two components: the SDF on a uniform grid and geometries that are irregular and vary for each sample. These points include adjacency relations, which can originate from the original grid. K-hop sampling is optional. For small-scale datasets, full-graph input without K-hop sampling can be utilized.}
		\label{fig:method overview}
	\end{figure}
	
	\subsection{Forward Pass Overview}\label{sec:overview}

	Let \(\mathbf{V} \in \mathbb{R}^{h \times w \times c}\) denote the uniform grid input (\(\mathbf{V} \in \mathbb{R}^{h \times w \times d \times c}\) for 3D situation), with \(h\), \(w\) and \(d\) as the grid dimensions and \(c\) the number of channels. Let \(\mathbf{X} = (\mathbf{X}_{\text{nodes}}, \mathbf{X}_{\text{edges}})\) represent the irregular graph input, where \(\mathbf{X}_{\text{nodes}} \in \mathbb{R}^{N \times c}\) are the node features, with \(N\) nodes, \(c\)-dimensional features and connectivity described by \(\mathbf{X}_{\text{edges}}\).
	
	The overall process of our proposed method is illustrated in Fig.\ref{fig:method overview}. Our method requires two inputs: the geometric SDF voxel (pixel for 2D case) and the geometric graph structure.  
	It can be roughly described as followed three steps:

	\paragraph{Grid Input Processing.} U-Net \(\mathcal{U}\) Process $\mathbf{V}$ through a U-Net $\mathcal{U}$ to extract features: $\mathbf{F}_{\mathbf{V}} = \mathcal{U}(\mathbf{V})$.  
	
	\paragraph{Graph Input Processing.} GNN \(\mathcal{G}\) process $\mathbf{X}$ through a graph neural network $\mathcal{G}$ to extract graph-based features: $\mathbf{F}_{\mathbf{X}} = \mathcal{G}(\mathbf{X})$. To address specific physical phenomena, domain experts often apply adaptive mesh refinement, resulting in dense subregions that encode prior knowledge of the underlying physical processes. We adopt K-hop sampling~\cite{nikolentzos2020k}, which enables learning from the original mesh while maintaining the local details. 
	
	\paragraph{Decoding for Final Prediction.} The extracted features from regular grid, $\mathbf{F}_{\mathbf{V}}$, are interpolated through spatial bilinear onto the vertices of the sampled unstructured mesh to obtain $\mathbf{F}_{\mathbf{V}}^{\mathcal{G}}$. $\mathbf{F}_{\mathbf{V}}^{\mathcal{G}}$ and $\mathbf{F}_{\mathbf{X}}$ are then concatenated and passed through a decoder network to generate the final prediction, $\hat{u} = \text{Decoder}(\mathbf{F}_{\mathbf{V}}^{\mathcal{G}}, \mathbf{F}_{\mathbf{X}})$.  In practice, We ensure that the U-Net and FVGN have identical hidden sizes and equal neural network depths (8-layers),  the Decoder is a three MLPs with the GeLU activation function.

	\subsection{Background Grid \& U-Net}\label{sec:attu}
	
	\begin{wrapfigure}{r}{8cm}
		\vspace{-0.5cm}
		\centering
		\includegraphics[width=0.43\textwidth]{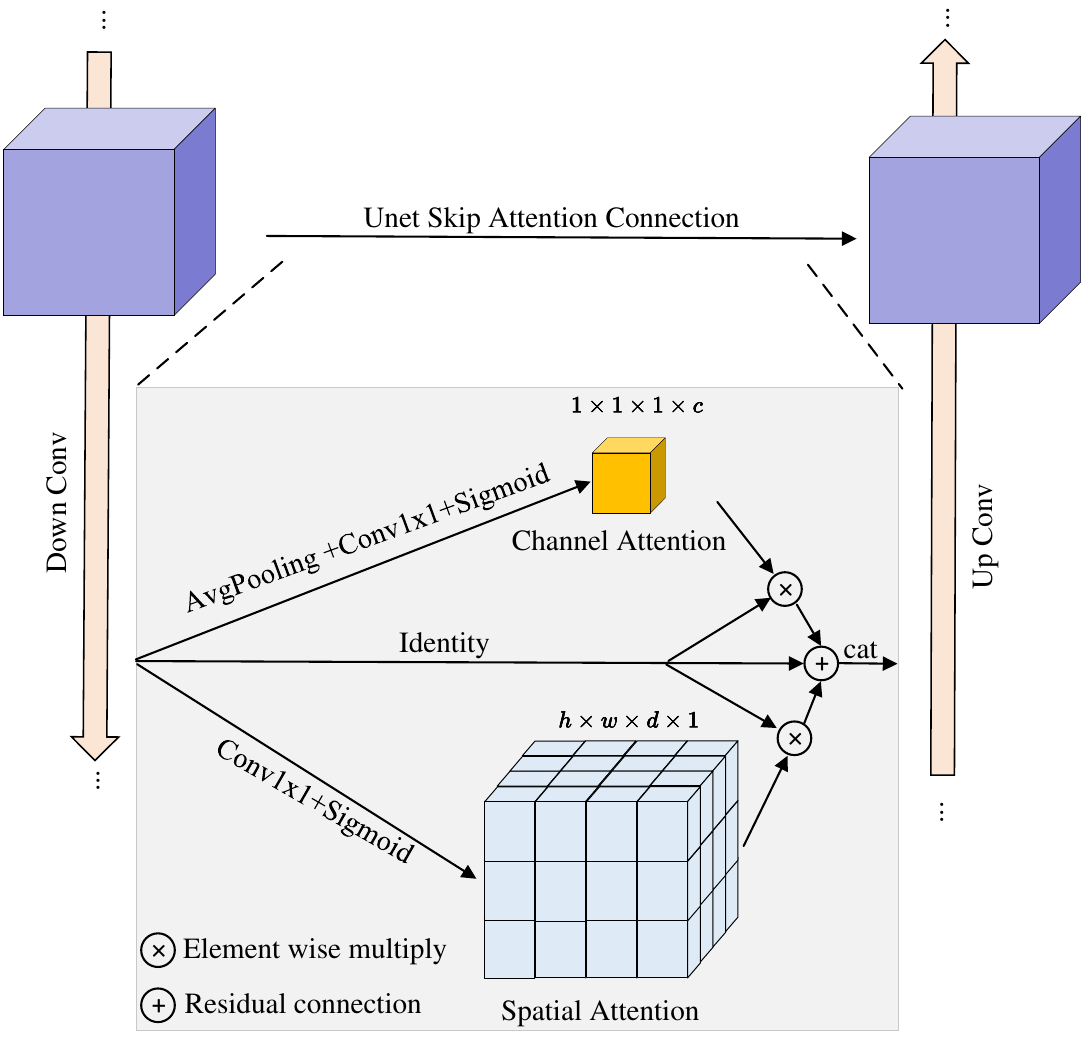}
		\caption{Attention skip connection (ASC).}
		\label{fig:unet}
		\vspace{-0.3cm}
	\end{wrapfigure}
	
	As illustrated in Fig.\ref{fig:bgm_sample}, the background grid is constructed as a uniform Cartesian grid that tightly encloses the geometric shape of the object. This grid aligns with the length, width, and height of the geometry, providing a consistent and regular discretization of the spatial domain. On this background grid, we compute the SDF of the geometric shape, resulting in SDF voxels that represent the distance from each grid point to the nearest surface of the geometry. To effectively process the SDF voxels and extract meaningful features, we employ a 3D convolutional neural network based on the U-Net architecture~\cite{islam2021braintumorsegmentationsurvival}.

	Specifically, as shown in Fig.\ref{fig:unet}, we use a 3D U-Net variant with Attention Skip Connections (ASC), combining channel and spatial attention. Channel attention computes weights via global average pooling, while spatial attention generates weights by reducing the channel dimension. Both are applied to the input through element-wise multiplication, summed, and concatenated with the upsampling layer input.
	
	
	\begin{equation}\label{eq:ASC_0}
		\text{Attn}_{ch}(\mathbf{F_v}^{\prime}) = \sigma(\phi_{ch}(\text{AvgPool}(\mathbf{F_v}^{\prime})))
	\end{equation}
	
	\begin{equation}\label{eq:ASC_1}
		\text{Attn}_{sp}(\mathbf{F_v}^{\prime}) = \sigma(W\mathbf{F_v}^{\prime} + b)
	\end{equation}
	
	\begin{equation}\label{eq:ASC}
		\text{ASC}(\mathbf{F_v}^{\prime}) = \mathbf{F_v}^{\prime} \odot \left(\text{Attn}_{ch}(\mathbf{F_v}^{\prime}) + \text{Attn}_{sp}(\mathbf{F_v}^{\prime}) + 1 \right)
	\end{equation}

	\paragraph{Multi-scale Contextual Modeling.} In our method, the U-Net possesses a hierarchical architecture, which effectively compensates for the difficulty that GNNs methods face in designing multi-scale frameworks. The background grid not only assists GNNs trained with K-hop sampling but also benefits Transolver~\cite{wu2024transolver} architectures based on randomly sampled point clouds. In subsequent experiments, we demonstrate the effectiveness of this method.

	\subsection{Node-based Finite Volume Graph Network}\label{ssec:fvgn}

	\textcolor{black}{
		Node-based Finite Volume Graph Network is a simplified variant of FVGN~\cite{li2024predicting}, which uses a two-step message aggregation GN block with extended message aggregation templates, with the cell block omitted. This modification is necessitated by the inherent complexity and heterogeneity of industrial CFD mesh cells (e.g., various combinations of triangles, tetrahedra, and polyhedra), which complicates consistent cell block calculation. Additionally, we employ directed edges to halve the number of latent edge representations, reducing GPU memory consumption. To mitigate potential overfitting caused by unidirectional edges, a random edge reversal is applied during training. These design choices allow the Node-based FVGN to achieve performance comparable to MGN with substantially lower memory usage. The subsequent text also refers to Node-based FVGN as FVGN for short.}
	
	The message-passing processing sequence of FVGN consists of \( M \) identical message passing layers, which encompass the Graph Network blocks~\cite{pfaff2021learning,sanchez2018graph}. The operations in the Node Block and Edge Block are defined by Eq.\eqref{FVGN-processor1} and Eq.\eqref{FVGN-processor2}, respectively.

	%
	%
	
	\begin{equation}\label{FVGN-processor1}
		\mathbf{s}_{i} \leftarrow \dfrac{1}{\vert \mathcal{N}_{i}\vert} \left(\sum_{j \in \mathcal{N}_{i}} {}  \mathbf{e}^{(l)}_{ij} \right); \mathbf{v}^{(l+1)}_i \leftarrow \phi^{vp}\left(\mathbf{v}^{(l)}_{i}, \dfrac{1}{\vert \mathcal{N}_i\vert} \sum_{j\in \mathcal{N}_i}^{} s_{j}\right)  
	\end{equation}

	\begin{equation}\label{FVGN-processor2}
		h_i \leftarrow \dfrac{1}{\vert \mathcal{N}_i \vert } \left(\sum_{j \in \mathcal{N}_{i}} {}  \mathbf{v}^{(l)}_{j} \right); 
		\mathbf{e}^{(l+1)}_{ij} \leftarrow \phi^{ep}\left(\mathbf{e}_{ij}^{(l)},h_{i}, h_{j}\right)
	\end{equation}

	In the Node Block (Eq.\eqref{FVGN-processor1}), FVGN first aggregates \(l\)-th layer edge features \(\mathbf{e}_{ij}^{(l)}\) to the shared nodes \(s_i\) and subsequently combines information from neighboring nodes \(s_j\) to the center node \(\mathbf{v}_{i}^{(l)}\). The original vertex features \(\mathbf{v}_{i}^{(l)}\) are concatenated with these aggregated features and updated via a Node-MLP \(\phi^{vp}\), which includes two hidden layers, GeLU activation functions, and LayerNorm for enhanced stability and performance. Node block output next layer vertex feature \(\mathbf{v}^{(l+1)}_{i}\).
	
	In the Edge Block (Eq.\eqref{FVGN-processor2}), FVGN aggregates features from neighboring nodes \(\mathbf{v}_j^{(l)}\) to the center nodes \(h_i\). Then, the features of the two vertices (\(h_i\), \(h_j\)) connected by an edge are concatenated with the edge feature \(\mathbf{e}_{ij}^{(l)} \) and updated using an Edge-MLP \(\phi^{ep}\), which has the same architecture as the Node-MLP. Edge block output next layer edge feature \(\mathbf{e}^{(l+1)}_{ij} \).

	In Eq.\eqref{FVGN-processor1} and Eq.\eqref{FVGN-processor2}, \( \mathcal{N}_i \) represents the vertices neighbour with vertex \(i\). Here, the message aggregation process's former places variables at the vertices.  In practice, we randomly sample K-hop subgraphs as inputs to this component for extracting mesh features. Learning directly from the original grid helps preserve detailed local information. Compared with MGN, FVGN achieves comparable performance with significantly lower GPU memory usage~\cite{li2024predicting} by only using unilateral edge. Additionally, we employ relative node features, ensuring that the neural network does not overfit to absolute coordinates or features but instead focuses on learning local patterns.

	\subsection{Loss function}
	
	The model was trained on three benchmarks using relative $L_2$ norm loss, addressing various aspects of flow field prediction, including pressure, velocity, and turbulence kinetic viscosity. 
	
	The Eq.\eqref{eq:loss_func} represents the relative $L_2$ norm loss for the flow field, where \(u\) denotes the predicted vector at each point, and \(u^{GT}\) is the ground truth. \(C\) represents the dimensionality of the solution space,\(B\) is the number of samples and \(N\) is the number of points per sample.
	\begin{equation}
		\label{eq:loss_func}
		\mathcal{L} = \frac{1}{\mathcal{B}} \sum_{i=1}^{\mathcal{B}} \frac{\sqrt{\sum_{j=1}^{N} \sum_{k=1}^{C} (u_{jk} - u^{GT}_{jk})^2}}{\sqrt{\sum_{j=1}^{N} \sum_{k=1}^{C} (u^{GT}_{jk})^2}}
	\end{equation}


	\section{Experiments}\label{sec:exp}

	In designing the experiments, we focus on MGN~\cite{pfaff2021learning}, FVGN~\cite{li2024predicting}, GINO~\cite{li2023geometryinformed}, and Transolver~\cite{wu2024transolver} as baselines, as they represent SOTA methods in geometry representation learning. Additionally, U-Net demonstrates strong performance in multi-scale modeling. We highlight the effectiveness of multi-scale modeling in these tasks by considering ASC-U-Net, which uses bilinearly interpolated outputs for decoding. GN-based methods such as MGN and FVGN are grounded in graph neural networks, while GINO, U-Net and Transolver are point-based approaches. Among all the baselines, Transolver demonstrated superior performance.

	\subsection{Datasets}

	\begin{table}
		\caption{Summary of the datasets. This table consists of the average point number of mesh, input, output, and data split. \(\boldsymbol{x}\) is the coordinate, $\mathbf{n}$ is the surface normal,$\mathbf{u}$ is the inlet velocity. \(\boldsymbol{v}, p,\nu_t\) represent the velocity, pressure, and turbulence viscosity, respectively.}
		\label{tab:dataset_detail}
		
		\centering

		\renewcommand{\multirowsetup}{\centering}
		
		\begin{sc}
			\begin{tabular}{l|c|c|c|c}
				\hline
				Benchmarks & \begin{tabular}[c]{@{}c@{}}Mesh Avg. \end{tabular} & Input & Output & Split \\ \hline
				ShapeNet-Car & 32,000 & $\boldsymbol{x}, SDF,\mathbf{n}$ & $\boldsymbol{v},p$ & (789, 100) \\
				AirfRANS & 180,000 & $\boldsymbol{x}, \mathbf{u}, SDF,\mathbf{n}$ & \begin{tabular}[c]{@{}c@{}}$\boldsymbol{v},p,\nu_t$\end{tabular} & (800, 200) \\
				DrivAerNet & 500,000 & $\boldsymbol{x}, SDF,\mathbf{n}$ & $p$ & (2772, 595) \\ \hline
			\end{tabular}
		\end{sc}

	\end{table}
	
	We use the ShapeNet-Car~\cite{umetani2018learning}, the AirfRANS~\cite{bonnet2022airfrans} and the DrivAerNet~\cite{elrefaie2024drivaernet}, which represent key examples of modern industrial design and are widely adopted for evaluating neural network performance. we show summary of these datasets in Tab.\ref{tab:dataset_detail}, a brief introduction to each dataset is provided below.
	
	\paragraph{ShapeNet-Car.} The ShapeNet-Car includes 889 samples with a diverse range of car shapes. Each sample contains the spatial velocity field on a volume mesh with about 32,000 elements, and the surface pressure on a surface mesh with about 3,700 elements.
	
	\paragraph{AirfRANS} The AirfRANS is a high-fidelity CFD dataset for studying the two-dimensional flow over airfoils at a subsonic regime. The dataset includes 1,000 samples, each containing the velocity, pressure, and turbulent kinematic viscosity fields on refined meshes.
	
	\paragraph{DrivAerNet.} The DrivAerNet is a large-scale, high-fidelity CFD dataset featuring 3D industry-standard car shapes. Each sample of this dataset comprises approximately 0.5 million surface meshes, offering high-resolution surface pressure and wall shear stress data, making it highly suitable for aerodynamic performance prediction.

	\subsection{Training Configuration}\label{ssec:config}
	For the model parameters, we employ an Attention U-Net with a depth of 4 and a channel list of [128, 256, 512, 1024], as well as an 8-layer FVGN with a hidden size of 128. For the baseline configuration, we fix the parameters of MGN and FVGN to 15 layers and a hidden size of 128, and the parameters of Transolver to 8 layers, a hidden size of 256, and 32 slices. For the ShapeNet-Car, the size of the points and edges in the original grid is manageable, so we use the full graph input without sampling. To balance performance and memory consumption, we randomly sample 2,000 subgraphs with a skip of k = 5 from the DrivAerNet and the AirfRANS (large-scale data) for training for graph based methods(MGN, FVGN, ME-GNN), and randomly sampled [1000, 3000, 9000, 16000, 32000] points and report the best score for points based methods(GINO, Transolver), we use a NVIDIA A100 40G to conduct all of the experiments. 
	
	We employ the Adam optimizer with a learning rate of $1e^{-3}$ and 4 batch size. Different numbers of epochs are required for the three datasets, We set [120, 600, 80] epochs for training on ShapeNet-Car, AirfRANS, and DrivAerNet respectively. The learning rate start with $1e^{-3}$ and reduce it to \(1e^{-4}\) for the last 5 epochs, starting 5 epochs before reaching the designated number of epochs.	
	
	\begin{figure}
		\centering
		\includegraphics[width=\textwidth]{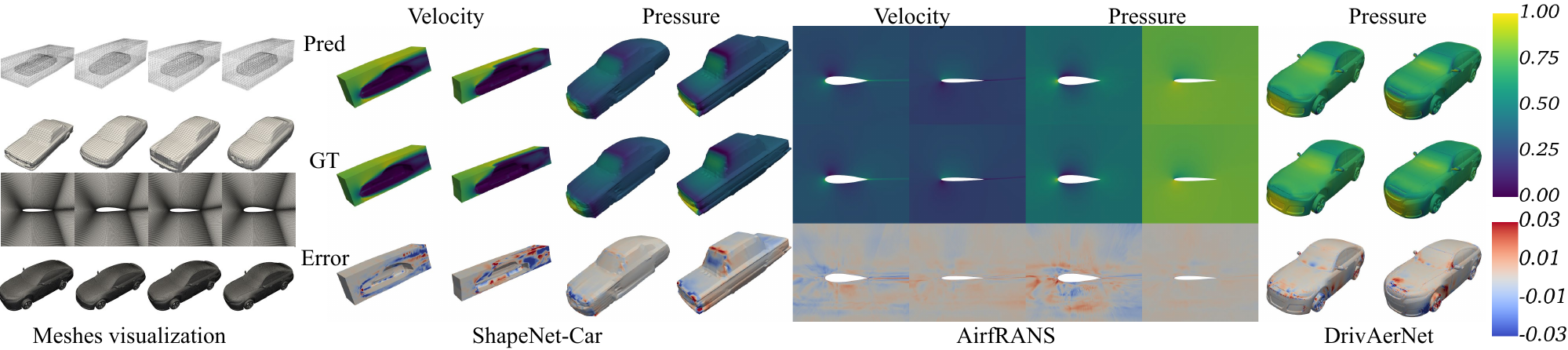}
		\caption{Visualization for the meshes and min-max normalized prediction of ME-GNN, ground truth, and error for the three datasets.}
		\label{fig:all_long}
	\end{figure}

	\subsection{Main Results}\label{ssec:result}
	\begin{table}
		\caption{Performance on the ShapeNet-Car and the AirfRANS, metrics include surrounding (Vol) and surface (Surf) physics field. For ShapeNet-Car, metrics is relative \( L_2 \) errors for Vol and Surf. For AirfRANS, metrics is MSE on normalized fields, the lift coefficient (\( C_L \)) is relative error, and Spearman’s correlation (\( \rho_L \)) for the lift coefficient. Baseline results are from Transolver\protect~\cite{wu2024transolver}. Bold highlights the best performance, while underlined text marks the second-best.}
		\label{tab:mainres_design}
		\vspace{-5pt}
		\vskip 0.15in
		\centering
		\begin{small}
				\renewcommand{\multirowsetup}{\centering}
				\setlength{\tabcolsep}{4.8pt}
				\begin{tabular}{l|cc|cccc}
					\hline
					\multicolumn{1}{l|}{\multirow{3}{*}{Model$^\ast$}} & \multicolumn{2}{c|}{ShapeNet-Car} & \multicolumn{4}{c}{AirfRANS} \\ \cline{2-7} 
					\multicolumn{1}{c|}{} & Vol $\downarrow$ & Surf $\downarrow$ & Vol $\downarrow$ & Surf $\downarrow$ & \multicolumn{1}{c|}{$C_{L}$ $\downarrow$} & $\rho_{L}$ $\uparrow$ \\
					\multicolumn{1}{c|}{} & \multicolumn{2}{c|}{$(\times 10^{-2})$} & \multicolumn{3}{c|}{$(\times 10^{-2})$} &  \\ \hline

					GNO & 3.83 & 8.15 & 2.69 & 4.05 & \multicolumn{1}{c|}{20.16} & 0.9938 \\
					
					Galerkin & 3.39 & 8.78 & 0.74 & 1.59 & \multicolumn{1}{c|}{23.36} & 0.9951 \\
					GNOT & 3.29 & 7.98 & 0.49 & 1.52 & \multicolumn{1}{c|}{19.92} & 0.9942 \\
					3D-GeoCA & 3.19 & 7.79 & / & / & \multicolumn{1}{c|}{/} & / \\
					\hline
					MGN & 3.54 & 7.81 & 2.14 & 3.87 & \multicolumn{1}{c|}{22.52} & 0.9945 \\
					GINO & 3.86 & 8.10 & 2.97 & 4.82 & \multicolumn{1}{c|}{18.21} & 0.9958 \\
					Transolver & 2.07 & 7.45 & \underline{0.37} & \underline{1.42} & \multicolumn{1}{c|}{\underline{10.30}} & \underline{0.9978} \\ 
					\hline
					FVGN & 4.60 & 11.48 & 3.36 & 4.87 & \multicolumn{1}{c|}{22.87} & 0.9923 \\
					ASC-U-Net(interp) & \textbf{1.80} & \underline{5.68} & 8.66 & 8.56 & \multicolumn{1}{c|}{17.91} &  0.9958\\
					ME-GNN(Ours) & \underline{1.96} & \textbf{5.56} & \textbf{0.33} & \textbf{0.67} & \multicolumn{1}{c|}{\textbf{3.72}} & \textbf{0.9993} \\ \hline
				\end{tabular}
			\end{small}
			\vspace{-5pt}
		\end{table}

			This section evaluates the performance of ME-GNN from several perspectives. Specifically, we visualize the prediction-ground truth error for three datasets(ShapeNet-Car, AirfRANS, and DrivAerNet) and provide an error comparison between our model and a set of baseline models. Quantitatively, the standard metric results for these datasets (ShapeNet-Car, AirfRANS, and DrivAerNet) are recorded. It should be noted that for the experiments on ShapeNet-Car and AirfRANS, we followed the configuration of Transolver~\cite{wu2024transolver}, and the baseline results are cited from this work.

		We visualize meshes and the velocity and pressure prediction of ME-GNN on the three datasets, as shown in Fig.\ref{fig:all_long}. After min-max normalization, the errors are generally below 0.03, demonstrating the robust performance of ME-GNN across various geometries and mesh types and densities.

		For the ShapeNet-Car, as shown in Tab.\ref{tab:mainres_design}, the performance gap between GNO and MGN is minimal, and the performance of FVGN is lower than that of MGN. Transolver achieves a significant lead over other models in the velocity field prediction. ME-GNN achieves a good overall performance among all models, with a relative \(L_2\) error of 0.0196 for the velocity field and 0.0556 for the surface pressure, representing improvements of 5\% and 25\% compared to Transolver, respectively. In this task, the ASC-U-Net achieved the performance(1.80 \(\times 10 ^{-2}\) and 5.58 \(\times 10 ^{-2}\) for volume amd surface pressure prediction, respectively), demonstrating the effectiveness of multi-scale neural networks in flow field prediction tasks. Due to the sparse meshes of the ShapeNet-Car, U-Net did not show disadvantages in such tasks, and the voxel resolution did not present a bottleneck.

		For the AirfRANS, as shown in Tab.\ref{tab:mainres_design}, the evaluation metrics for volume and surface are MSE on normed fields. Self-attention-based models such as Galerkin, GNOT, and Transolver outperform graph-based methods like MGN, FVGN and GNO. Among these, Transolver performs best, achieving an MSE of 0.0037 for the volume field, 0.0142 for surface pressure, and a relative error of 0.0103 for the lift coefficient. In this task, the boundary layer near the airfoil is extremely dense, and U-Net faces a performance bottleneck caused by the interpolated pixels method. ME-GNN outperforms Transolver in volume field prediction with an MSE of 0.0033, representing an 8\% improvement. It also achieves a 52\% reduction in surface pressure MSE, with a value of 0.0067, and a 54\% improvement in lift prediction, yielding a relative error of 0.0372 (a 63.8\% improvement). ME-GNN significantly outperforms U-Net on AirfRANS, demonstrating GNN’s superior adaptability to unstructured data.

		\begin{wraptable}{r}{8.3cm}
			\caption{Metrics on the DrivAerNet. Relative $L_2$ of the surface (Surf) pressure and the relative error of drag coefficient ($C_{D}$) are recorded, along with their Spearman’s rank correlations $\rho_{D}$.}
				\begin{tabular}{l|ccc}
					\hline
					Model\textbackslash{}Metrics       & Surf$\downarrow$ &  $C_D$  $\downarrow$ & $\rho_D$ $\uparrow$ \\ \hline
					MGN       & 0.2394                       & 0.0696                       & 0.6464              \\
					GINO & 0.1858                       & 0.0614                       & 0.7941              \\
					Transolver~ & \underline{0.1441}                       & \underline{0.0239}           & \underline{0.9632}    \\ \hline
					FVGN       & 0.1978                       & 0.0687                       & 0.7362              \\
					ASC-U-Net(interp)      & 0.1872                       & 0.0247                       & 0.9597              \\
					ME-GNN (Ours)            & \textbf{0.1416}              & \textbf{0.0231}              & \textbf{0.9655}     \\ \hline
				\end{tabular}
			\label{tb:TrackC_metrics}
			\vspace{-0.2cm}
		\end{wraptable}

		For the DrivAerNet, as shown in Tab.\ref{tb:TrackC_metrics}, ME-GNN achieved a surface pressure relative \(L_2\) error of 0.1416 and a relative error for correlation coefficient of 0.9655, outperforming MGN, FVGN, GINO, and Transolver. The relative error of FVGN is lower than that of MGN, suggesting the advantage of FVGN over MGN on high-resolution meshes.

\begin{figure}
			\centering
			\includegraphics[width=0.9\textwidth]{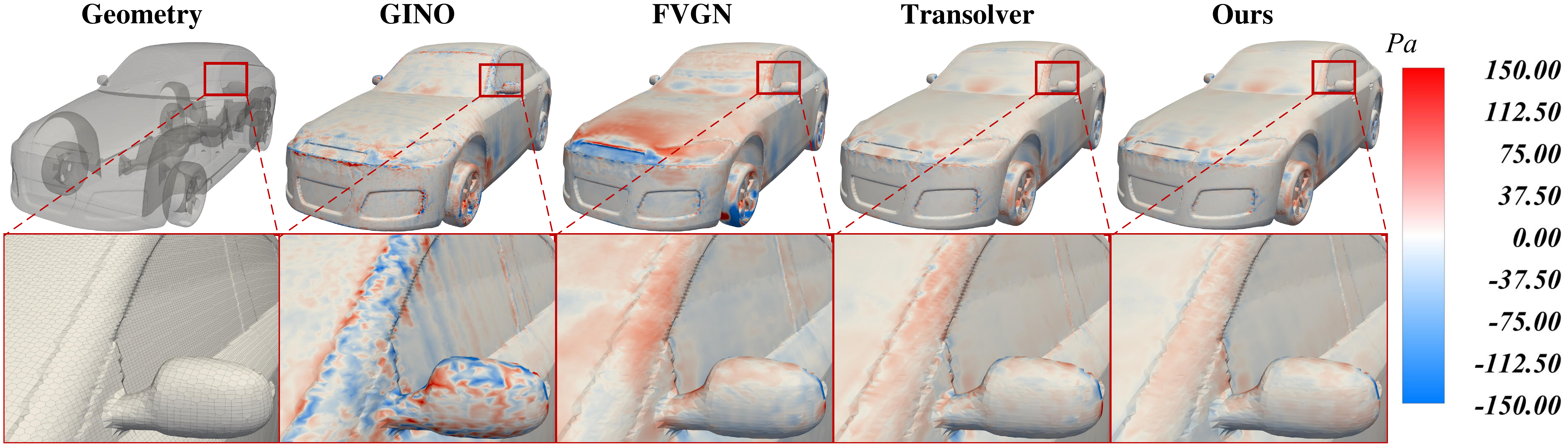}
			\caption{Pressure prediction error visualization.}
			\label{fig:driv_error_viz}
		\end{figure}

		The error visualization comparison, illustrated in Fig.\ref{fig:driv_error_viz}, provides additional insights into the model's performance. GINO, which relies on global frequency-domain representations, demonstrates fewer overall errors than FVGN; however, it exhibits higher errors in capturing local fine-grained geometric details, such as the vehicle's side mirror. On the other hand, FVGN shows lower errors in specific localized areas, such as the side mirror; however, it has significant inaccuracies on the engine hood and front windshield, where the flow field experiences large-scale variation trend, and capturing the global context is essential. Transolver also produced competitive results. Overall, ME-GNN achieves the lowest errors in fine-grained regions. It suggests that ME-GNN's ability to learn from the detailed surface mesh has enabled it to retain local features while effectively integrating strong global representations.

		Across all three datasets, ME-GNN demonstrated consistent superiority over FVGN, GINO, and other baseline models. By overcoming the limitations of grid discretization and leveraging multi-scale feature extraction, ME-GNN effectively captured fine-grained geometric details and integrated global context, resulting in state-of-the-art predictive performance. These results validate ME-GNN’s potential to address complex aerodynamic prediction tasks with improved accuracy and efficiency.

		\section{Ablation Studies}\label{sec:abla study}

		We conducted ablation experiments using the DrivAerNet and AifRANS. The analysis focused on the impact of number and method of sampling, number of K-hop, network width, and background grid resolution on the neural network's performance and efficiency. 
		
		\begin{figure}
			
			\centering
			
			\begin{subfigure}{0.45\textwidth}
				\centering
				\includegraphics[width=\textwidth]{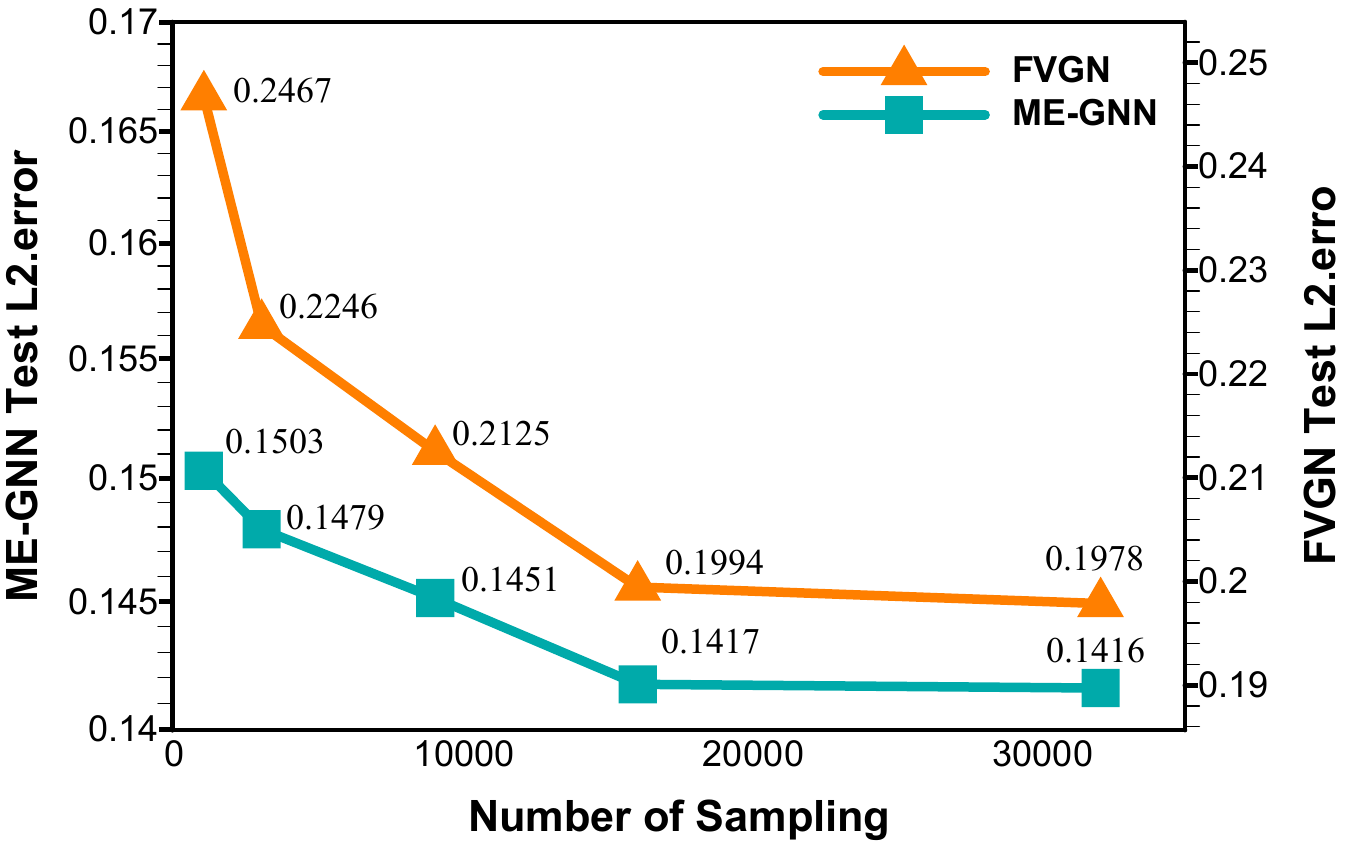}
				\caption{Varying sampling for ME-GNN.}
				\label{fig:sampling_fvgn}
			\end{subfigure}
			\begin{subfigure}{0.45\textwidth}
				\centering
				\includegraphics[ width=\textwidth]{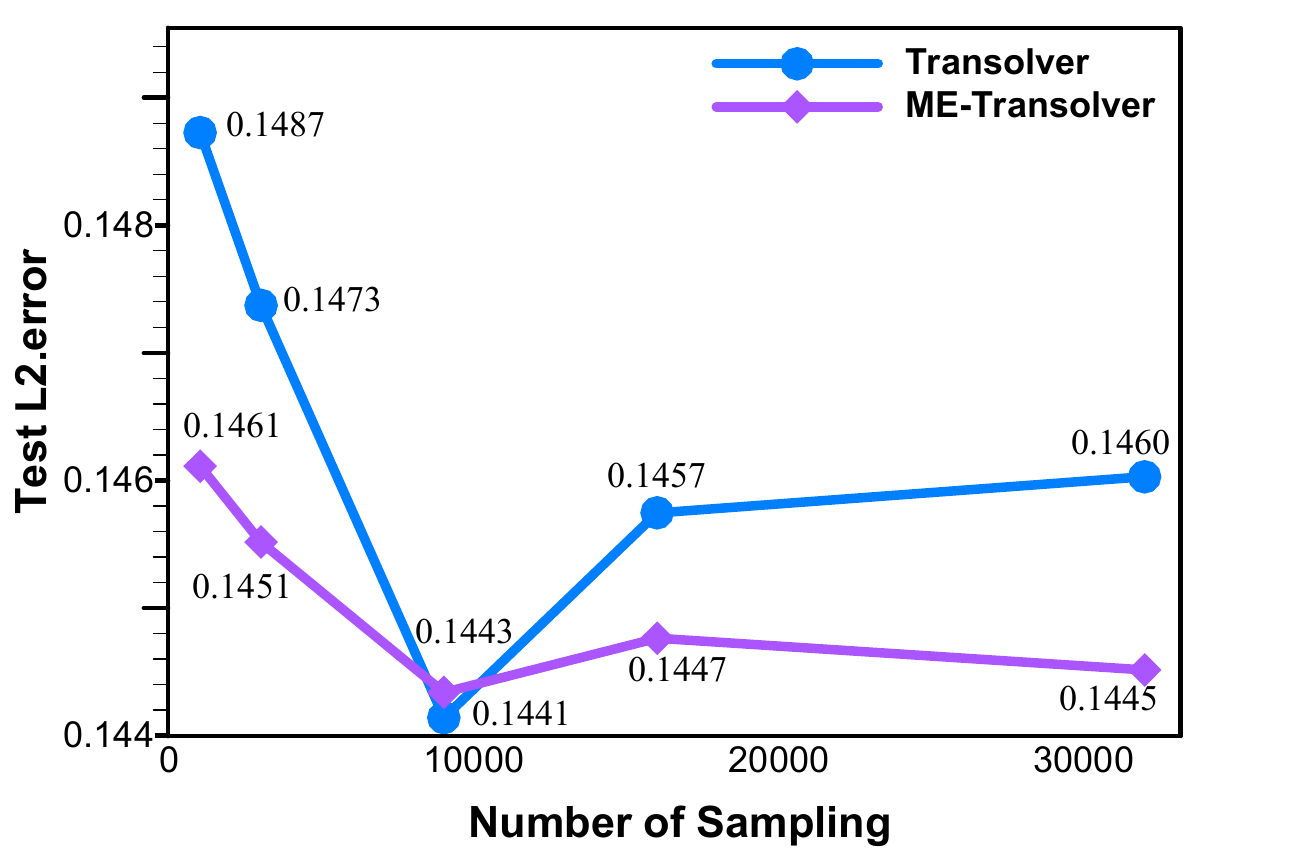}
				\caption{Varying sampling for ME-Transolver.}
				\label{fig:sampling_transo}
			\end{subfigure}
			
			\caption{Ablation study for number of sampling.}
			\label{fig:abla_sampling}
		\end{figure}

		\subsection{Sampling Ablation}

		Given the strong performance of U-Net on this task, we replace the part of FVGN with Transolver (ME-Transolver) and reducing the hidden size to 128 for comparison. For the graph-based method, we control the number of subgraphs sampled and fix the 5-hop to maintain approximately 1000, 3000, 9000, 16000, and 32000 sampling nodes for DrivAerNet.

		The sampling rate impact on model performance for FVGN and ME-GNN is shown in Fig.\ref{fig:sampling_fvgn}. Under small sampling point settings, with about 1000 and 3000 nodes for sampled subgraphs, multi-scale feature enhancement significantly improves FVGN, reducing the error from 0.2467 and 0.2246 to 0.1503 and 0.1479, respectively. This substantially reduces the overall error of the graph neural network and addresses the issue of long-distance modeling dependence in graph neural networks. GNN performance is impacted by the sampling size, as the sample size significantly influences the completeness of the constructed graph relative to the original data. For Transolver, as shown in Fig.\ref{fig:sampling_transo}, which possesses global representation capabilities, the enhancement also provides improvement under the same small sampling, reducing the error from 0.1487 and 0.1473 to 0.1461 and 0.1451, respectively. The results underscore the effectiveness of using multi-scale feature enhancement.

		\begin{wrapfigure}{r}{8cm}
			
			\centering
			\includegraphics[ width=0.45\textwidth]{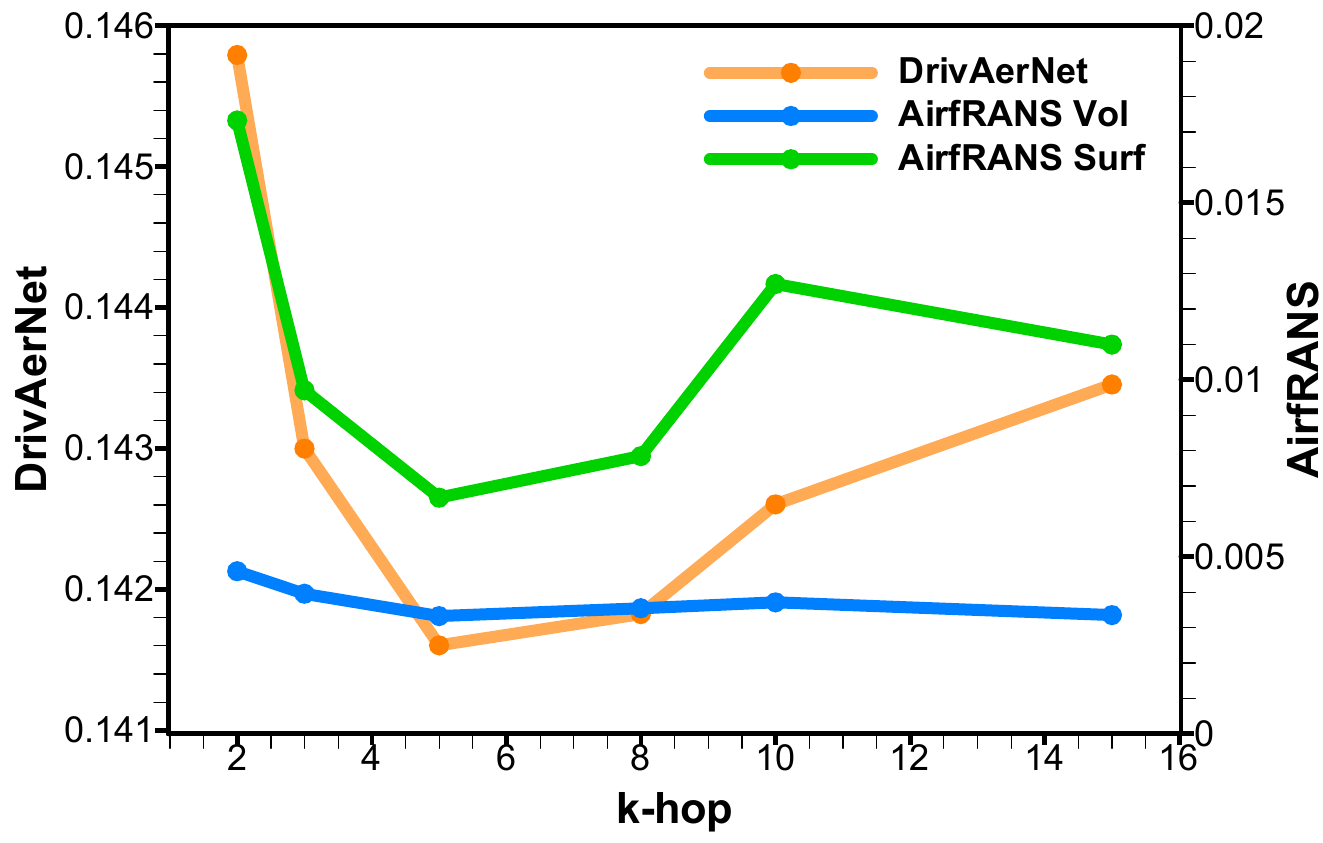}
			\caption{Different k hop}
			\label{fig:khop_abla}
			\vspace{-0.2cm}
		\end{wrapfigure}

			Secondly, we conducted an ablation experiment on different K-hop subgraph sampling quantities. To eliminate the impact of the total number of sampled points, we controlled the total number of points to be around 3,2000 by adjusting the number of subgraphs sampled, as shown in Fig.\ref{fig:khop_abla}. The lowest surface pressure error on both the DrivAerNet and AirfRANS datasets was achieved when $k = 5$. The choice of $k$ in the $k$-hop subgraph sampling notably influences surface pressure prediction, whereas its impact on flow field prediction is comparatively minor.  This could be because the flow field in the boundary region and the boundary layer undergoes drastic changes. When \( k \) is too small, it fails to capture effective local geometric surface features. Conversely, when \( k \) is too large, it can lead to an overly sparse spatial distribution of samples. In tasks involving external flow fields and geometric surfaces (such as AirfRANS), this may reduce the likelihood of sampling the geometric boundaries effectively.
			
					\begin{wraptable}{r}{8.3cm}
				\caption{Ablation experiment with alternative sampling method.}
					
					\begin{tabular}{l|ccc}
						\hline
						Sampling & \multicolumn{2}{c}{AirfRANS} & DrivAerNet \\ \hline
						\multirow{2}{*}{Metrics} & Vol & Surf & Surf \\
						& \multicolumn{2}{c}{MSE (\(\times10^{-2}\))} & Relative L2 \\ \hline
						Random \& kNN & 2.34 & 3.26 & 0.1612 \\
						Random \& Radius graph & 2.21 & 3.19 & 0.1603 \\
						
						K-hop & \textbf{0.33} & \textbf{0.67} & \textbf{0.1416} \\ \hline
					\end{tabular}

					\label{tb:khop_abla}
					\vspace{-0.2cm}
				\end{wraptable}

			
				To evaluate the importance of the sampling strategy, we replaced the K-hop method with an alternative approach. In this setup, we randomly sampled 32,000 points and subsequently constructed either a kNN graph or a radius graph, with a maximum of 16 neighbors per node.
			
				As shown in the corresponding table, this change resulted in a significant performance degradation. For the AirfRANS dataset, using a kNN graph increased the volumetric and surface errors from 0.33 and 0.67 to 2.34 and 3.26, respectively; using a radius graph yielded a similar increase to 2.21 and 3.19. For the DrivAerNet dataset, the error rose from 0.1416 to 0.1612 with the kNN graph and to 0.1603 with the radius graph.
			
				This marked decline in accuracy underscores the crucial role of K-hop sampling. Its ability to preserve the local mesh topology is evidently vital for effective representation learning in mesh-based GNNs.

			It is worth noting that the performance degradation of AirfRANS is much greater than that of DrivAerNet. This discrepancy arises because AirfRANS includes a very dense boundary layer mesh, whereas DrivAerNet consists only of a surface mesh. A properly configured U-Net is effective at learning surface geometry representations from the SDF (as shown in Tab.~\ref{tb:TrackC_metrics}, based on the U-Net's prediction results). However, when it comes to learning representations of CFD spatial discretization meshes, the U-Net struggles with the complexity of dense boundary layer meshes. Moreover, during inference, the kNN edges generated by ME-GNN’s sampling process maintain consistent topological relationships with the training data. Nonetheless, the distribution of edge features deviates, leading to a notable drop in performance for the graph network, while the impact on DrivAerNet (which uses only surface meshes) remains minimal.

			\subsection{Model Ablation}

						\begin{table}
				\centering
				\caption{Ablation experiment for FVGN.}
				\begin{tabular}{c|ccc|cc}
					\hline
					\multirow{3}{*}{Model(interp)$^\ast$} & \multicolumn{3}{c|}{AirfRANS} & \multicolumn{2}{c}{DrivAerNet} \\ \cline{2-6} 
					& Vol $\downarrow$ & Surf $\downarrow$ & Memory $\downarrow$ & Surf $\downarrow$ & $C_{D}$ $\downarrow$ \\
					& \multicolumn{2}{c}{$(\times 10^{-2})$} & GB & \multicolumn{2}{c}{} \\ \hline
					MGN(origin) & 7.16 & 8.33 & 14.3 & 0.2394 & 0.0696 \\
					MGN & 10.76 & 9.83 & 8.8 & 0.3237 & 0.1172 \\
					FVGN-Full & 8.66 & 8.56 & 8.9 & 0.1978 & 0.0687 \\ 
					FVGN-Node & 9.32 & 9.41 & 8.7 & 0.1994 & 0.0696 \\
					FVGN-Edge & 9.88 & 9.70 & 8.8 & 0.2043 & 0.0712 \\ \hline
				\end{tabular}
				
				\label{tb:fvgn_abla}
			\end{table}
			
				This subsection presents a detailed ablation study to evaluate the contribution of the proposed modules and the sensitivity to key hyper-parameters. Specifically, we investigate the impact of the node-based FVGN configuration, the hidden size, and the latent grid resolution.
			
				An ablation study was conducted to evaluate the node-based components of FVGN on the AirfRANS and DrivAerNet datasets. All models were trained with a batch size of 1, using 2,000 points sampled from each data sample (approximately 30,000 points total) and a K-hop neighborhood of k=5. Since FVGN extends MeshGraphNets (MGN), our analysis focused on comparing five model configurations.
			
				The first configuration (MGN origin), which serves as a baseline, is the original MGN architecture with bi-directional edges. This design doubles the number of latent edge features and consequently increases GPU memory requirements. The remaining four configurations employ unique directed edges with a random reversal operation applied during training. These variants include: FVGN-Node, which incorporates only the FVGN node update process, and FVGN-Edge, which utilizes only the FVGN edge update process. In both variants, all other architectural components revert to the standard MGN design.
			
				The tab.\ref{tb:fvgn_abla} present a comparative analysis of FVGN against MGN. A key observation is that transitioning the standard MGN from resource-intensive bi-directional edges to memory-efficient uni-directional edges causes a significant degradation in prediction accuracy; for instance, on the DrivAerNet dataset, the surface error increases from 0.2394 to 0.3237. Our primary finding is that FVGN successfully addresses this performance gap. Despite using uni-directional edges and consuming substantially less GPU memory (8.9 GB vs. 14.3 GB for the original), the full FVGN model achieves performance comparable to, and in the case of DrivAerNet, superior to the original bi-directional MGN (achieving a surface error of 0.1978 vs. 0.2394).

				Crucially, the ablation study confirms the utility of FVGN's individual components. Both the FVGN-Node and FVGN-Edge variants substantially outperform the standard uni-directional MGN. On the DrivAerNet dataset, for example, they reduce the surface error to 0.1994 and 0.2043, respectively, from the baseline of 0.3237 for the uni-directional MGN. This demonstrates that these modules are effective in recovering the performance lost from abandoning bi-directional edges. We speculate that this performance advantage stems from FVGN's extended message-passing template, which provides a larger receptive field than the first-order message passing in MGN, enabling it to better capture complex flow phenomena.
			
\begin{figure}
				
				\centering
				
				\begin{subfigure}{0.47\textwidth}
					
					\centering
					\includegraphics[width=\textwidth]{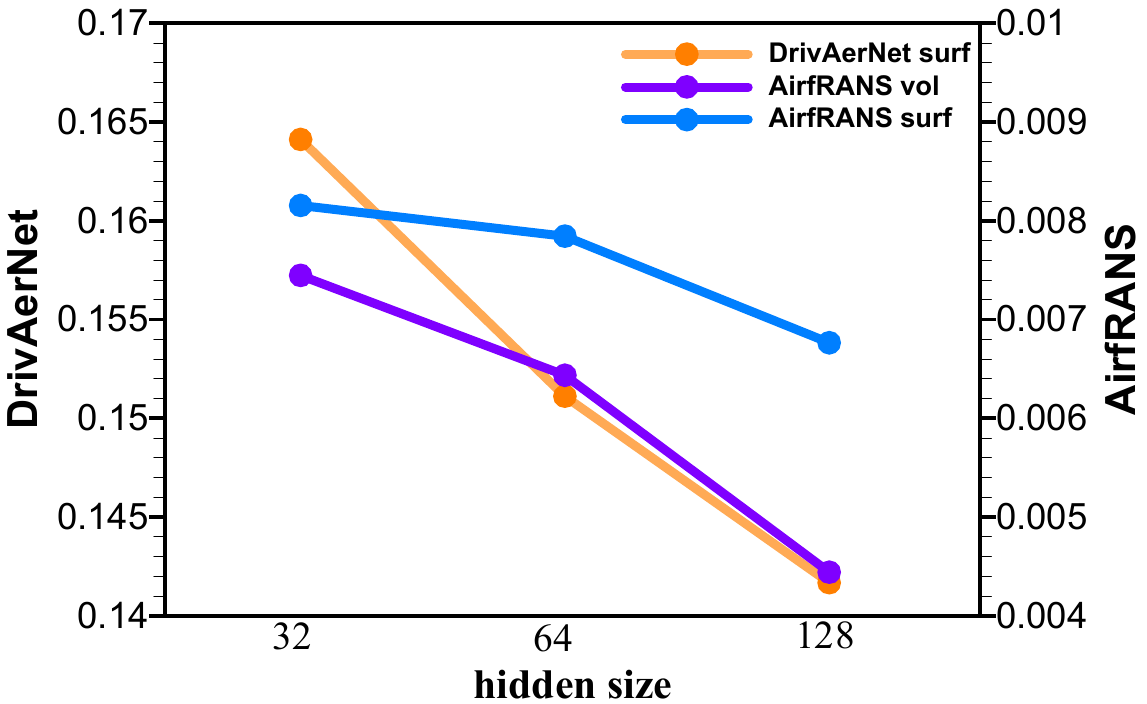}
					\caption{Effect of hidden size.}
					\label{fig:hidden}
				\end{subfigure}
				\begin{subfigure}{0.45\textwidth}
					\centering
					\includegraphics[width=\textwidth]{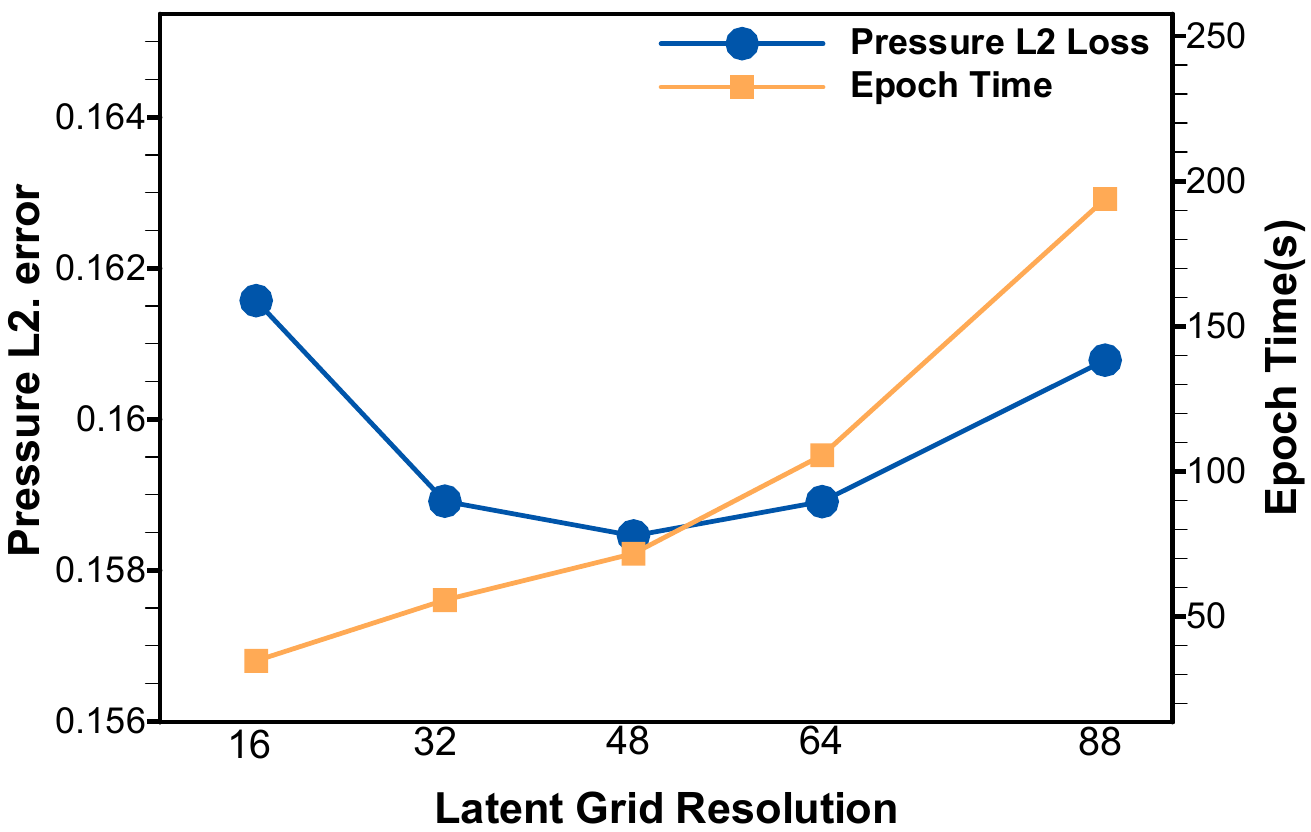}
					\caption{Effect of uniform grid resolution.}
					\label{fig:voxel_ref}
				\end{subfigure}
				
				\caption{Model ablation.}
				\label{fig:model_abla}
			\end{figure}

			%
			We analyze the impact of hidden size and background grid resolution on ME-GNN. As shown in Fig.\ref{fig:hidden}, we conducted ablation experiments on the ME-GNN's hidden size, using values of 32, 64, and 128 on the DrivAerNet and AirfRANS datasets. The results show that the ME-GNN's performance improves as the hidden size increases, demonstrating the scalability of ME-GNN.

			We utilize 500 DrivAerNet samples for training and 50 samples for evaluation. As shown in Fig.\ref{fig:voxel_ref}, a grid resolution of approximately 48 was chosen as it achieves an optimal balance between computational cost and performance, accounting for the cubic growth of the voxel grid. In the future, we can adopt more efficient methods to handle the voxel data, allowing it to have greater scalability in terms of both data and hidden size.

				%
		%

		\subsection{Efficiency analysis}\label{sec:eccicency_analy}

		\begin{figure}
			
			\centering
			
			\begin{subfigure}{0.45\textwidth}
				\centering
				\includegraphics[width=\textwidth]{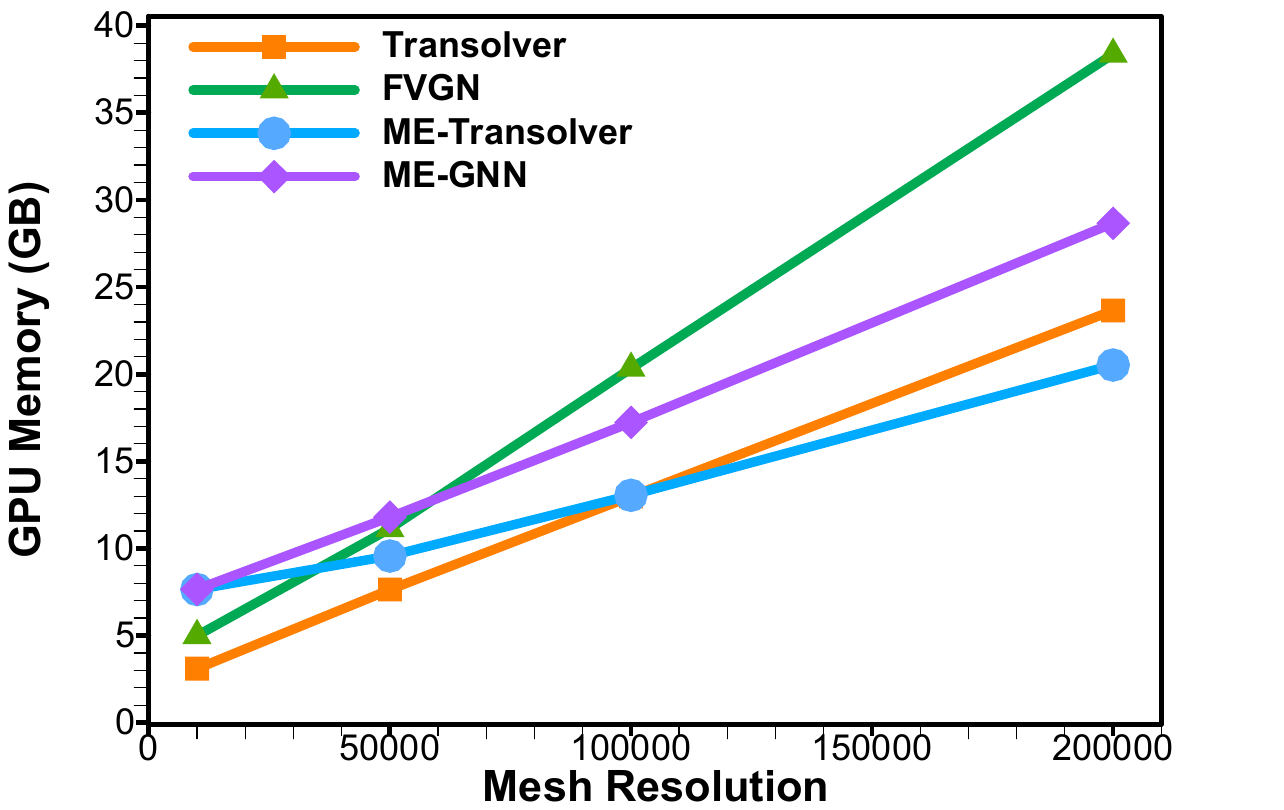}
				\caption{GPU memory occupation.}
				\label{fig:abla_gpu_mem}
			\end{subfigure}
			\begin{subfigure}{0.45\textwidth}
				\centering
				\includegraphics[width=\textwidth]{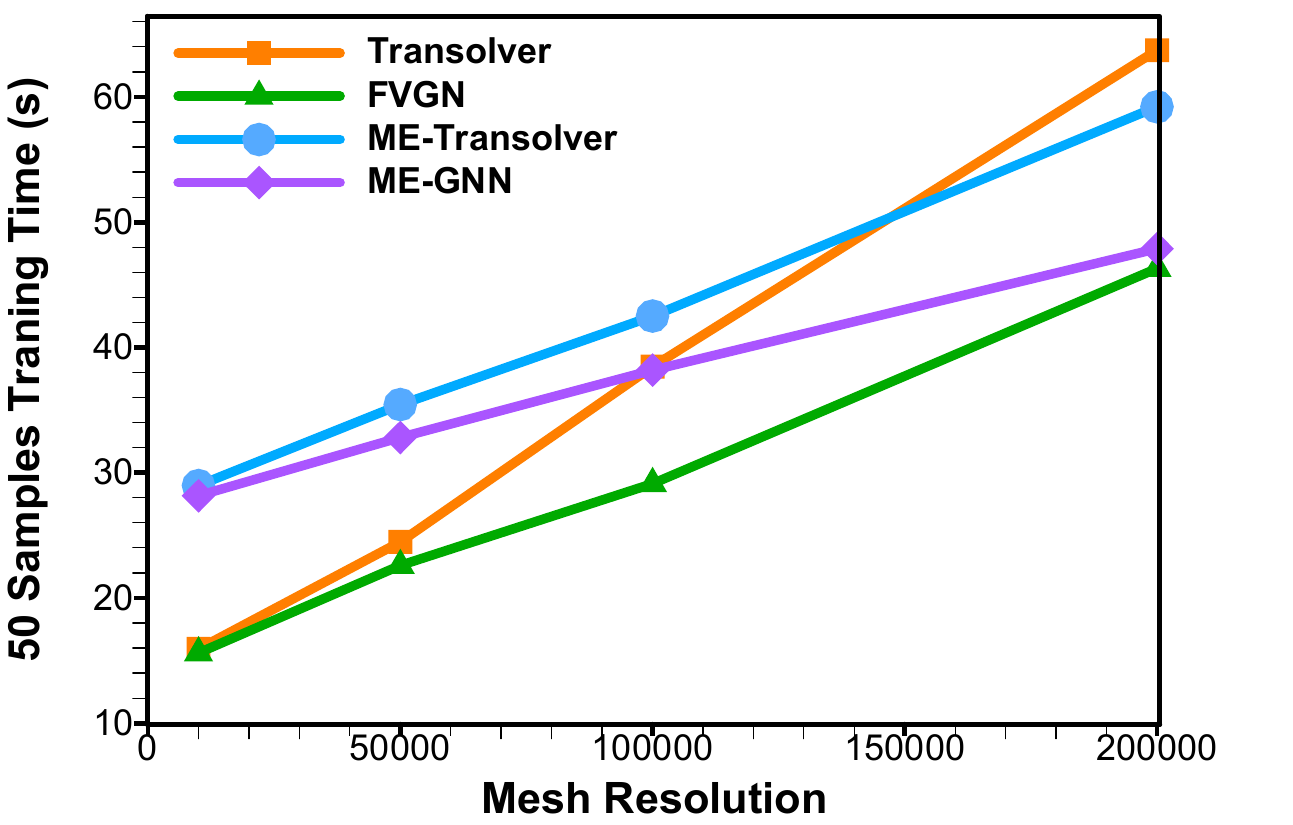}
				\caption{50 samples training epoch time.}
				\label{fig:abla_time}
			\end{subfigure}

			\caption{Efficiency comparison.}
			\label{fig:efficiency-results}
		\end{figure}

		Theoretically, the computational complexity of ME-GNN is \( O(N + E + V) \), where \( N \), \( E \), and \( V \) represent the number of mesh points, edges, and background uniform grid elements, respectively. This is a linear time complexity. Transolver also has a linear time complexity of \( O(NM + M^2) \), demonstrating impressive performance, where \( M \) is the slice number. Using the DrivAerNet as an example, for a mesh, the ratio of the number of points to edges is approximately 1:2 (undirected edges, no duplicates). We used this ratio for input and trained 50 samples to test GPU memory usage and time consumption. Visualizations were conducted with input values of [1, 5, 10, 20] \(\times 10^4\). The results show that (Fig.\ref{fig:abla_gpu_mem} and \ref{fig:abla_time}), in terms of GPU memory usage, the consumption and growth rates of ME-GNN and Transolver are similar. In terms of computation time, ME-GNN outperforms Transolver after a mesh resolution of \( 10 \times 10^4 \), with a lower growth rate. Due to the shallower graph network layers, ME-GNN also demonstrates better efficiency compared to FVGN.

		As shown in the Fig.\ref{fig:efficiency-results}, ME-GNN achieves linear time complexity, and its growth rate is relatively smooth, which makes it suitable for tasks on large-scale datasets. However, currently, the voxelized or pixelized U-Net results in higher initial computational resource and memory requirements.

		\section{Conclusion}\label{sec:dis_and_concl}
		
		This paper presents a hybrid U-Net and GNN architecture for fluid dynamics prediction, leveraging a uniform SDF grid and detailed surface mesh to process geometric information separately. The U-Net is responsible for extracting multi-scale geometric features, while the GNN captures local details. By integrating both fine and coarse features, this architecture enhances prediction accuracy.

		We evaluate ME-GNN on three benchmark datasets. It shows excellent performance in predicting velocity and pressure fields on the ShapeNet-Car, achieving relative \(L_2\) errors of 0.0196 and 0.0556, respectively. Additionally, ME-GNN performed exceptionally well in aerodynamic performance prediction for datasets with dense meshes(AirfRANS, DrivAerNet). It achieved a relative error of 0.0372 for the lift coefficient on the AirfRANS and 0.0231 for the drag coefficient on the DrivAerNet. 
		
		We utilize advanced CNN-based U-Net methods to address sampling limitations, reducing the performance degradation of neural networks in configurations with small samples. This approach is highly suitable for industry applications with constrained sampling budgets, where enhancing efficiency and minimizing costs is crucial.
		

		
		\bibliographystyle{cas-model2-names}
		
		\bibliography{cas-refs}
		
		
		
	\end{document}